\documentclass{article}

\usepackage[preprint]{neurips_2025}

\usepackage[utf8]{inputenc} 
\usepackage[T1]{fontenc}
\usepackage{hyperref}
\usepackage{url}
\usepackage{booktabs}
\usepackage{amsfonts}
\usepackage{nicefrac}
\usepackage{microtype}
\usepackage{xcolor}

\usepackage{soul}
\usepackage{url}
\usepackage{textcomp}
\usepackage[small]{caption}
\usepackage{graphicx}
\usepackage{amsmath}
\usepackage{amsthm}
\usepackage{booktabs}
\usepackage{algorithm}
\usepackage{algorithmic}
\usepackage[switch]{lineno}
\usepackage{color}
\usepackage{multirow}
\usepackage{adjustbox}
\usepackage{booktabs}
\usepackage{multirow}
\usepackage{graphicx}
\usepackage{amssymb}
\usepackage{pifont}
\usepackage{threeparttable}
\usepackage{array}
\usepackage{xcolor}
\usepackage{color, colortbl}
\usepackage{tabularx}
\usepackage{booktabs}
\usepackage{ragged2e}  
\usepackage{enumitem}

\usepackage{array}
\usepackage{booktabs}
\usepackage{multirow}
\usepackage{bxcoloremoji}

\newcolumntype{C}[1]{>{\centering\arraybackslash}m{#1}}
\newcolumntype{P}[1]{>{\raggedright\arraybackslash}p{#1}}
\newcommand{\shifth}[1]{\makebox[0.52cm][c]{#1}}
\newcommand{\taskh}[1]{\makebox[0.80cm][c]{#1}}
\newcommand{\taskcell}[1]{\makebox[0.80cm][c]{#1}}
\newcommand{\cmark}{\ding{51}}
\definecolor{Gray}{gray}{0.9}

\newtheorem{definition}{Definition}

\newcolumntype{M}[1]{>{\raggedright\arraybackslash}m{#1}}
\newcolumntype{C}[1]{>{\centering\arraybackslash}m{#1}}

\title{Out-of-Distribution Generalization in Graph Foundation Models}

\author{%
Haoyang Li,\;
Haibo Chen,\;
Xin Wang\thanks{Corresponding authors},\;
\textbf{Wenwu Zhu}\footnotemark[1]\\
Tsinghua University\\
\small \texttt{lihy218@gmail.com},\;
\small \texttt{chb24@mails.tsinghua.edu.cn},\;
\small \texttt{\{xin\_wang, wwzhu\}@tsinghua.edu.cn}\\
}

\begin{document}

\maketitle

\begin{abstract}
Graphs are a fundamental data structure for representing relational information in domains such as social networks, molecular systems, and knowledge graphs. However, graph learning models often suffer from limited generalization when applied beyond their training distributions. In practice, distribution shifts may arise from changes in graph structure, domain semantics, available modalities, or task formulations. To address these challenges, graph foundation models (GFMs) have recently emerged, aiming to learn general-purpose representations through large-scale pretraining across diverse graphs and tasks. In this survey, we review recent progress on GFMs from the perspective of out-of-distribution (OOD) generalization. We first discuss the main challenges posed by distribution shifts in graph learning and outline a unified problem setting.  We then organize existing approaches based on whether they are designed to operate under a fixed task specification or to support generalization across heterogeneous task formulations, and summarize the corresponding OOD handling strategies and pretraining objectives. Finally, we review common evaluation protocols and discuss open directions for future research. To the best of our knowledge, this paper is the first survey for OOD generalization in GFMs.
\end{abstract}

\section{Introduction}

Graphs are a fundamental data structure for representing relational information in many applications, including social and information networks, molecular and biological systems, recommendation platforms, and knowledge graphs~\cite{xu2018powerful,wu2022graph,li2023curriculum}. By encoding entities as nodes and interactions as edges, graphs capture complex dependency structures that are difficult to model using independent feature representations. Graph learning methods, such as graph neural networks, have become a central tool for predictive and reasoning tasks, including node classification, link prediction, and graph-level prediction~\cite{wu2020comprehensive,li2021intention}. However, models trained on a specific dataset or graph often exhibit limited generalization when applied to new testing environments, where graph topology, feature distributions, semantic meaning, or task formulation differ from those seen during training~\cite{li2022ood}.

Out-of-distribution (OOD) generalization provides a useful perspective for addressing these limitations~\cite{li2025out,wang2025uncertainty}. In the field of graph learning, distribution shifts may arise from multiple sources. Structural properties such as connectivity patterns or motif statistics can change across graphs~\cite{li2022learning,zhang2024disentangled}. Domain-specific factors, including data collection and annotation practices, may introduce dataset biases~\cite{hu2020open}. Auxiliary modalities, such as text or molecular features, may be missing, noisy, or inconsistently aligned across datasets~\cite{cai2024multimodal}. In addition, downstream tasks may change in supervision form or output space~\cite{pasini2025multi}. These factors make it difficult to deploy graph models reliably under in-the-wild environment with diverse distribution shifts.

Recently, graph foundation models (GFMs) have emerged and attracted growing attention from the research community. Inspired by foundation models in language and vision~\cite{bommasani2021opportunities,zhou2025comprehensive}, GFMs aim to learn general-purpose graph representations through large-scale pretraining on diverse graph collections~\cite{liu2025graph,wang2025graph}. Instead of only optimizing for a specific dataset, these models explore capturing generalizable patterns that can be reused and stable across graphs, domains, and downstream objectives. An increasing number of work has explored different approaches to building GFMs~\cite{sun2023all,wang2024gft,yuanmuch}, including multi-graph pretraining~\cite{lachi2024graphfm}, alignment across domains and modalities~\cite{he2025unigraph2}, invariant representation learning~\cite{wangtowards2025},  prompt- or instruction-based interfaces for task generalization~\cite{he2025unigraph}, etc. There foundation models address both practical and methodological limitations of traditional graph learning. In many applications, labeled data for new graphs is scarce, and retraining models from scratch is often infeasible~\cite{li2021disentangled}. Distribution shifts across domains and time further make stable deployment difficult. From a modeling perspective, the heterogeneity of graph structures and feature spaces also bring challenges to the design of architectures and objectives that scale across different scenarios~\cite{hu2020open}. Pretraining on large and diverse graph data provides a way to improve sample efficiency, enable transfer, and reduce reliance on dataset-specific correlations, motivating increased interest in GFMs~\cite{zhang2024can,xia2024anygraph,wangtowards2025}, providing a promising paradigm for handling distribution shifts for OOD generalization.

Several recent surveys~\cite{li2023survey,liu2025graph,wang2025graph} have reviewed GFMs from perspectives such as model architecture~\cite{wang2025modular}, pretraining objectives, scalability, and application domains~\cite{wu2025graph,yuan2025graver}. In contrast, this survey organizes the literature explicitly from the perspective of OOD generalization. Rather than focusing on model design alone, we examine how different GFMs address distribution shifts arising from changes in graph structure, domain semantics, modality availability, and task formulation, providing a deeper discussion on existing methods and clarifying the relationship between modeling choices, deployment challenges, and evaluation practices.

In this survey, we provide a comprehensive overview of graph foundation models from the perspective of OOD generalization. We first identify the key challenges posed by distribution shifts in graph learning and introduce a unified problem formulation that captures the OOD in structure/feature, domain, modality, and task. We then organize existing methods into two broad categories according to whether they explicitly support generalization across different task specifications. The first category includes approaches that focus on generalization under a fixed task setting, where OOD generalization is achieved by learning representations that remain effective across structural, domain, or modality shifts. The second category comprises methods that are designed to generalize across more complex heterogeneous tasks. Within each category, we analyze representative methods. We also review the widely-adopted evaluation settings used to validate OOD performance and discuss open challenges and future directions. Last but not least, we discuss potential future research topics, which could shed light on the development of this promising topic. We hope that this survey provides useful insights for promoting research in the community.

\section{Challenges and Problem Formulation}

GFMs can support learning and inference across diverse environments, where the data-generating process may differ substantially between training and deployment. In such settings, the poor generalizations are often driven by systematic distribution shifts. In this section, we first identify the core challenges that arise when applying graph foundation models under distribution shift. We then formalize OOD generalization in GFMs through a unified problem formulation.

\subsection{Challenges}

OOD generalization in GFMs involves challenges spanning the \textbf{structural}, \textbf{domain}, \textbf{modality}, and \textbf{task} levels. These challenges reflect the difficulty of learning representations that are both broadly transferable and generalizable to distributional shifts. We summarize four core challenges below.

\begin{itemize}[leftmargin = 0.5cm]

\item \textbf{Structural-level: Structural diversity and unstable graph semantics.}  
Graphs differ widely in topology, node attributes, and relational patterns across settings. Spurious correlations that hold only under specific structural regimes further harm generalization. Although alignment and invariance-based objectives have been proposed~\cite{zhu2025graphclip,wang2025multi,yu2025samgpt}, extracting stable and transferable structural representations remains difficult for GFMs.

\item \textbf{Domain-level: Knowledge integration without interference.}  
Pretraining on multiple graph datasets can improve coverage but often induces negative transfer due to conflicting structural or semantic patterns. Shared models may become biased toward dominant domains or suppress domain-specific signals. Techniques such as mixture-of-experts and adaptive routing~\cite{xia2024anygraph,tang2025graphmoe} partially alleviate this issue, but stable multi-domain integration remains challenging.

\item \textbf{Modality-level: Inconsistent cross-modal signals and alignment.}  
Graphs are frequently augmented with auxiliary modalities such as text or molecular features, whose availability and quality vary across datasets. Models risk over-reliance on certain modalities and degrade when modalities are missing or shifted. While alignment and gating mechanisms improve integration~\cite{he2025unigraph2,wang2025towards}, generalization under modality shift remains limited.

\item \textbf{Task-level: Generalization across diverse task formulations.}  
GFMs are expected to support heterogeneous downstream tasks with different output spaces and supervision granularity. Uniform generalization across tasks is difficult, and task-specific fine-tuning often leads to forgetting. Prompting and instruction-based methods offer partial solutions~\cite{liu2024one,he2025unigraph}, but task-agnostic remains a challenging problem.

\end{itemize}

\subsection{Problem Formulation}
\label{sec:definition}

The definition of OOD generalization in GFMs can be formulated as follows:

\begin{definition}[OOD Generalization in Graph Foundation Models]
\label{prob:definition}
Let $G=(V,E,X,M)\in\mathcal{G}$ denote a graph with structure $(V,E)$, node attributes $X$, and optional modalities $M$, and let $Y\in\mathcal{Y}$ denote the output; graphs are drawn from environments $e\in\mathcal{E}$ inducing distributions $p_e(G,Y)$, with training following the mixture $p_{\mathrm{src}}(G,Y)=\sum_{e\in\mathcal{E}_{\mathrm{src}}}\pi_e p_e(G,Y)$ where $\pi_e\ge 0$ and $\sum_{e\in\mathcal{E}_{\mathrm{src}}}\pi_e=1$, and testing following $p_{\mathrm{tgt}}(G,Y)$ with distribution shift $p_{\mathrm{src}}\neq p_{\mathrm{tgt}}$. The goal of OOD generalization in GFMs is to learn an optimal graph predictor $f^*_\theta$ that achieves minimal expected loss on the test distribution
\begin{equation}
f^*_\theta \;=\; \arg\min_{f\in\mathcal{F}} \; \mathbb{E}_{(G,Y)\sim p_{\mathrm{tgt}}}\!\left[\mathcal{L}(f(G),Y)\right],
\end{equation}
while model parameters $\theta$ are optimized using samples drawn from $p_{\mathrm{src}}$ and $\mathcal{L}$ is a loss function. Distribution shifts between $p_{\mathrm{src}}$ and $p_{\mathrm{tgt}}$ are induced by changes in latent generative factors $\Phi=(\Phi_{\mathrm{struct}},\Phi_{\mathrm{dom}},\Phi_{\mathrm{mod}},\Phi_{\mathrm{task}})$ governing $p_e$, where $\Phi_{\mathrm{struct}}$ controls the generative process of node attributes, graph topology and relational semantics (e.g., degree distributions, motif frequencies, or edge meaning), $\Phi_{\mathrm{dom}}$ captures domain-specific data collection and annotation mechanisms that induce dataset biases and spurious correlations, $\Phi_{\mathrm{mod}}$ governs the presence, quality, and alignment of auxiliary modalities, and $\Phi_{\mathrm{task}}$ determines the form, granularity, and semantics of supervision and output task spaces. OOD generalization corresponds to learning $f_\theta$ that is stable under distribution shifts in $\Phi$ between $\mathcal{E}_{\mathrm{src}}$ and $\mathcal{E}_{\mathrm{tgt}}$.

\end{definition}

Intuitively, each environment~\cite{arjovsky2019invariant,wang2022generalizing} corresponds to a particular configuration of graph structure/node feature, domain semantics, available modalities, and task formulation, as captured by the latent factors $\Phi$~\cite{li2025disentangling}. Structural shifts ($\Phi_{\mathrm{struct}}$) change node feature, graph topology and relational patterns, causing correlations learned in one graph family to break in another; domain shifts ($\Phi_{\mathrm{dom}}$) introduce dataset-specific biases that can dominate training and lead to negative transfer; modality shifts ($\Phi_{\mathrm{mod}}$) alter which signals are present or reliable, making models sensitive to missing or shifted modalities; and task shifts ($\Phi_{\mathrm{task}}$) change supervision types and output spaces, challenging uniform generalization across heterogeneous objectives. All these shifts arise from changes in the underlying data-generating process, and OOD generalization therefore requires GFM to avoid overfitting to any particular configuration of these factors and instead learn stable and generalizable mechanisms that remain predictive when structure, domain, modality, or task changes, enabling reliable transfer to previously unseen environments.

\begin{table*}[t]
\centering
\caption{Overview of Graph Foundation Models for OOD Generalization}
\label{tab:gfm_ood}
\begin{threeparttable}
\begin{adjustbox}{max width=0.999\textwidth}
\setlength{\tabcolsep}{0pt}
\begin{tabular}{
l 
@{\hspace{12pt}}C{1.0cm}@{\hspace{20pt}} 
C{0.65cm}@{\hspace{8pt}}C{0.65cm}@{\hspace{8pt}}C{0.65cm}@{\hspace{8pt}}C{0.65cm}@{\hspace{12pt}} 
P{5.2cm}@{\hspace{12pt}} 
P{4.5cm}@{\hspace{12pt}} 
C{0.80cm}@{\hspace{10pt}}C{0.80cm}@{\hspace{10pt}}C{0.93cm}
}
\toprule
\multirow{2}{*}{\scalebox{1.15}{\textbf{Model}}} & 
\multirow{2}{*}{\scalebox{1.15}{\textbf{Venue}}} & 
\multicolumn{4}{c}{\scalebox{1.15}{\textbf{Distribution Shifts}}} & 
\multirow{2}{*}{\scalebox{1.15}{\textbf{OOD Handling Strategy}}} & 
\multirow{2}{*}{\scalebox{1.15}{\textbf{Pretrain Paradigm}}} & 
\multicolumn{3}{c}{\scalebox{1.15}{\textbf{Downstream Tasks}}} \\
\cmidrule(lr){3-6} \cmidrule(lr){9-11}
& & \shifth{Str.} & \shifth{Dom.} & \shifth{Mod.} & \shifth{Tsk.} & & & \taskh{Node} & \taskh{Link} & \taskh{Graph} \\
\midrule
\rowcolor{Gray}
\multicolumn{11}{c}{\scalebox{1.1}{\textbf{Homogeneous-Task GFMs for OOD Generalization}}} \\
\midrule

GraphFM~\cite{lachi2024graphfm} & arXiv & \cmark & \cmark & & & Multi-graph diversity training & Multi-graph training & \taskcell{\cmark} & \taskcell{} & \taskcell{} \\
AnyGraph~\cite{xia2024anygraph} & arXiv & \cmark & \cmark &  & & MoE-based heterogeneity routing & Contrastive learning & \taskcell{\cmark} & \taskcell{\cmark} & \taskcell{\cmark} \\
MDGPT~\cite{yu2024text} & arXiv & \cmark & \cmark & & & Domain alignment & Contrastive learning & \taskcell{\cmark} & \taskcell{\cmark} & \taskcell{\cmark} \\
PatchNet~\cite{sun2025handling} & KDD & \cmark & \cmark & & & Learnable graph patching & Multi-graph pretraining & \taskcell{\cmark} & \taskcell{} & \taskcell{\cmark} \\
GraphAny~\cite{zhao2025graphany} & ICLR & \cmark & \cmark & & & Invariant entropy attention & Node label reconstruction & \taskcell{\cmark} & \taskcell{} & \taskcell{} \\
GraphLoRA~\cite{yang2025graphlora} & KDD & \cmark & \cmark & & & Low-rank structural alignment & Contrastive learning & \taskcell{\cmark} & \taskcell{} & \taskcell{} \\
MDGFM~\cite{wang2025multi} & ICML & \cmark & \cmark & & & Topology-aligned invariant learning & Contrastive learning & \taskcell{\cmark} & \taskcell{} & \taskcell{} \\
SAMGPT~\cite{yu2025samgpt} & WWW & \cmark & \cmark & & & Structural distribution adaptation & Contrastive learning & \taskcell{\cmark} & \taskcell{} & \taskcell{\cmark} \\
GraphCLIP~\cite{zhu2025graphclip} & WWW & \cmark & \cmark & & & Invariant contrastive alignment & Contrastive learning & \taskcell{\cmark} & \taskcell{\cmark} & \taskcell{} \\
RiemannGFM~\cite{sun2025riemanngfm} & WWW & \cmark & \cmark & & & Structure-invariant vocabulary & Contrastive learning & \taskcell{\cmark} & \taskcell{\cmark} & \taskcell{} \\
MDP-GNN~\cite{lin2025unified} & AAAI & \cmark & \cmark & & & Meta-domain alignment & Intra-/inter-domain alignment & \taskcell{\cmark} & \taskcell{} & \taskcell{} \\
GraphPFN~\cite{eremeev2025graphpfn} & arXiv & \cmark & \cmark & & & In-context Bayesian approximation & PFN loss + masked modeling & \taskcell{\cmark} & \taskcell{} & \taskcell{} \\
GOODFormer~\cite{liao2026invariant} & KDD & \cmark & \cmark & & & Invariant subgraph modeling & Invariant risk minimization & \taskcell{} & \taskcell{} & \taskcell{\cmark} \\
\midrule
\rowcolor{Gray}
\multicolumn{11}{c}{\scalebox{1.1}{\textbf{Heterogeneous-Task GFMs for OOD Generalization}}} \\
\midrule
OFA~\cite{liu2024one} & ICLR & \cmark & \cmark & & \cmark & Unified semantic prompting & Unified prompted classification & \taskcell{\cmark} & \taskcell{\cmark} & \taskcell{\cmark} \\
LLaGA~\cite{chen2024llaga} & ICML & \cmark & \cmark & & \cmark & Structure--semantic alignment & QA likelihood maximization & \taskcell{\cmark} & \taskcell{\cmark} & \taskcell{} \\
OpenGraph~\cite{xia2024opengraph} & EMNLP & \cmark & \cmark & & \cmark & Universal graph tokenization & Masked modeling & \taskcell{\cmark} & \taskcell{\cmark} & \taskcell{} \\
GOFA~\cite{kong2025gofa} & ICLR & \cmark & \cmark & & \cmark & Unified pretraining & Next-token prediction & \taskcell{\cmark} & \taskcell{\cmark} & \taskcell{\cmark} \\
LLM-BP~\cite{wang2025generalization} & ICML & \cmark & \cmark & & \cmark & Task-adaptive embeddings & Task-adaptive inference & \taskcell{\cmark} & \taskcell{\cmark} & \taskcell{} \\
GIT~\cite{wangtowards2025} & ICML & \cmark & \cmark & & \cmark & Invariant semantic pretraining & Task-tree reconstruction & \taskcell{\cmark} & \taskcell{\cmark} & \taskcell{\cmark} \\
GFT~\cite{wang2024gft} & NeurIPS & \cmark & \cmark &  & \cmark &
Transferable tree vocabulary &
Task-tree reconstruction &
\taskcell{\cmark} & \taskcell{\cmark} & \taskcell{\cmark} \\
AutoGFM~\cite{chen2025autogfm} & ICML & \cmark & \cmark & & \cmark & Invariant architecture customization & Architecture-aware contrastive & \taskcell{\cmark} & \taskcell{\cmark} & \taskcell{\cmark} \\
UniGraph~\cite{he2025unigraph} & KDD & \cmark & \cmark & & \cmark & Text-unified representation & Masked modeling & \taskcell{\cmark} & \taskcell{\cmark} & \taskcell{\cmark} \\
UniGraph2~\cite{he2025unigraph2} & WWW & \cmark & \cmark & \cmark & \cmark & MoE-based alignment & Masked modeling & \taskcell{\cmark} & \taskcell{\cmark} & \taskcell{\cmark} \\
\bottomrule
\end{tabular}
\end{adjustbox}
\begin{tablenotes}
\small
\begin{minipage}{0.95\textwidth}
\item \textbf{Distribution Shifts:} Str. = Structural (including topology and node attribute); Dom. = Domain; Mod. = Modality; Tsk. = Task.
\item \textbf{Downstream Tasks:} Node/Link/Graph indicate node-, link-, and graph-level tasks supported/evaluated in the original paper.
\end{minipage}
\end{tablenotes}
\end{threeparttable}
\end{table*}

\section{Categorization}
Building on the problem formulation in Definition~\ref{prob:definition}, we categorize existing GFMs for OOD generalization according to how they parameterize the prediction function $f_\theta$ under shifts in the latent generative factors $\Phi = (\Phi_{\text{struct}}, \Phi_{\text{dom}}, \Phi_{\text{mod}}, \Phi_{\text{task}})$. 
While all methods are trained on samples drawn from a source mixture distribution $p_{\text{src}}$, they differ fundamentally in whether the task specification $\Phi_{\text{task}}$ is assumed to be fixed or allowed to vary across environments. 
This distinction directly affects how models encode invariance, transfer supervision, and adapt representations when deployed under unseen target distributions $p_{\text{tgt}}$.
Accordingly, we organize the literature into two broad categories: (i) \textit{homogeneous-task} GFMs, which explore a fixed task formulation during pretraining and inference, and (ii) \textit{heterogeneous-task} GFMs, which explicitly model task variability or aim to generalize across multiple supervision types and output spaces.

\subsection{Homogeneous-Task GFMs for Handling OOD}
Homogeneous-task GFMs focus on settings where a single downstream task is fixed throughout pretraining and deployment, such as node classification, link prediction, or graph-level prediction. Under this single-task regime, distribution shifts arise from variations in graph structure, domain semantics, or auxiliary modalities, while the supervision form and output space remain unchanged. Models are therefore trained to optimize one task-specific objective across multiple source environments, with the goal of learning representations that remain stable under changes in $\Phi_{\text{struct}}$, $\Phi_{\text{dom}}$, and $\Phi_{\text{mod}}$. OOD generalization in this category is achieved by enforcing structural invariance, domain alignment, or modality-generalized encoding within a shared task formulation, rather than by adapting to new forms of supervision. Representative approaches emphasize multi-graph pretraining, contrastive alignment, invariant substructure modeling, or Bayesian-style in-context inference, while maintaining a consistent output interface across environments.

GraphFM~\cite{lachi2024graphfm} presents a multi-graph pretraining paradigm based on a Perceiver-style encoder that maps graphs of arbitrary size into a fixed set of latent tokens, forming a shared representation space. Node features are encoded and aggregated into the latent space through cross-attention and self-attention, enabling unified processing of graphs with diverse structures. It is pretrained on over one hundred graph datasets using a consistent node classification objective, while adaptation updates only lightweight dataset-specific components. By leveraging large-scale multi-graph diversity during pretraining, GraphFM encourages latent tokens to capture patterns that remain stable across domains, supporting generalization to unseen graphs.

Building on the idea of multi-graph coverage, AnyGraph~\cite{xia2024anygraph} introduces a mixture-of-experts foundation architecture to explicitly manage heterogeneity across graph datasets. AnyGraph trains multiple expert models and dynamically routes each graph to suitable experts based on self-supervised link prediction signals. A unified structure and feature embedding is constructed using singular value decomposition~\cite{wall2003singular} and simplified message passing, allowing graphs with different feature dimensions and adjacency sizes to share a common input space. Pretraining mixes graphs from many domains using link prediction as the main part of the objective. Therefore, AnyGraph handles distribution shifts through expert specialization and routing rather than global parameter adaptation. 

MDGPT~\cite{yu2024text} further develops the multi-domain pretraining by introducing explicit domain alignment mechanisms within a single shared encoder. Node features from each source domain are first projected into a shared latent space using domain-specific alignment functions, while learnable domain tokens modulate these features to preserve domain-dependent patterns. A universal link prediction objective is used to pretrain the encoder across domains, ensuring a consistent task interface. Therefore, OOD generalization is achieved through dual prompts that modulate representations without updating backbone parameters. It addresses domain-level shifts by separating shared semantic structure from domain-specific variation through token-based modulation, and by transferring knowledge via prompt combinations, enabling generalization to new unseen domains.

PatchNet~\cite{sun2025handling} introduces learnable graph patches to support unified pretraining across datasets with heterogeneous node features. It unfolds node attributes into fixed-size token channels and learns a patch structure that organizes these tokens into multiple patches guided by the original graph topology, enabling graphs from different domains to be encoded in a shared patch space. Patch-level representations are aggregated through a transformer-style module to produce node embeddings. It treats feature heterogeneity as a primary source of domain shift in cross-graph deployment. By learning patch construction and patch-level encoding during multi-graph pretraining, the model provides a mechanism to reduce feature mismatch across domains and support generalization to unseen graphs with distribution shifts.

GraphAny~\cite{zhao2025graphany} departs from parameter-sharing encoders and instead defines a fully inductive foundation model for node classification based on analytical predictors. It replaces trained GNN layers with a family of LinearGNNs whose parameters are computed in closed form using labeled nodes, making predictions independent of feature dimensionality and label identity. Multiple LinearGNNs with different graph filters are combined through an attention mechanism that operates on distances between prediction distributions. An entropy normalization step aligns these distances across graphs with different label cardinalities. It can well handle shifts in structure, feature space, and label space by construction, since both base predictors and attention inputs are permutation invariant and dimension agnostic.

GraphLoRA~\cite{yang2025graphlora} follows a pretrain-and-adapt foundation model design and introduces low-rank adaptation to support cross-graph generalization. A pretrained GNN backbone is frozen, while a parallel low-rank network captures task-specific updates. Feature mismatch is addressed through a structure-aware maximum mean discrepancy objective that aligns feature distributions while weighting nodes by graph diffusion, and structural differences are handled using contrastive learning between frozen and adapted representations. Because GraphLoRA explicitly models feature and structural discrepancies between source and target graphs and integrates them into a unified optimization process, it can well support efficient adaptation with a small number of trainable parameters for OOD generalization.

MDGFM~\cite{wang2025multi} extends multi-domain modeling by jointly aligning feature semantics and graph topology. Heterogeneous node features are projected into a shared semantic space using domain tokens, while topology alignment is achieved through structure refinement that balances node attributes and neighborhood information. Pretraining maximizes mutual information between original and refined graphs using a contrastive objective. Adaptation relies on meta and task prompts rather than backbone updates. In OOD settings, MDGFM addresses shifts in both structure and domain by explicitly aligning graphs toward shared semantic and topological representations, supporting transfer to unseen domains within a homogeneous task framework.

SAMGPT~\cite{yu2025samgpt} also adopts a text-free multi-domain foundation design but emphasizes structural adaptation through domain-specific structure tokens. A shared graph encoder is pretrained using contrastive objectives after aligning feature dimensions, while structure tokens modulate aggregation to capture domain-dependent patterns. During adaptation, holistic and specific prompts adjust structure-aware aggregation without modifying encoder parameters. This method treats cross-domain generalization as a problem of structural distribution adaptation, transferring knowledge by reweighting learned structure tokens rather than retraining for OOD generalization.

GraphCLIP~\cite{zhu2025graphclip} targets text-attributed graphs and constructs a foundation model through contrastive alignment between graph representations and language-based graph summaries. A graph transformer encodes sampled subgraphs, while summaries generated by large language models provide semantic anchors. Pretraining aligns graph and text representations in a shared space and is augmented with invariant alignment under graph perturbations. At deployment, text prompts enable direct inference on new graphs, with optional prompt tuning for low-data settings. For tackling OOD problem, GraphCLIP encourages representations that remain predictive across domain and structural changes by grounding graph semantics in language and enforcing consistency across perturbed environments.

RiemannGFM~\cite{sun2025riemanngfm} focuses on learning a structure-centered GFM based on a shared structural vocabulary. It maps tree-like and cyclic substructures into hyperbolic and spherical geometries, respectively, and combines them through a Riemannian product bundle. Universal Riemannian layers learn substructure embeddings and aggregate them into global representations. Pretraining uses geometric contrastive objectives without textual signals or augmentations. Under OOD settings, RiemannGFM supports generalization by learning geometry-aligned structural primitives that recur across graphs, enabling generalization across domains with different attributes and connectivity patterns.

MDP-GNN~\cite{lin2025unified} formulates multi-domain graph learning as alignment toward a latent meta-domain that captures shared structure and semantics. It integrates heterogeneous node features, learns inter-domain connections through pivot nodes, and aligns graph distributions using Wasserstein distance, all within a unified pretraining objective. By explicitly modeling domain-level shifts by mapping graphs into a shared latent space, it can support generalization to unseen domains through distribution alignment.

GraphPFN~\cite{eremeev2025graphpfn} adapts the prior-data fitted network paradigm to graphs by combining a tabular foundation model backbone with attention-based graph adapters. Global attention enables in-context learning from labeled nodes, while local adapters capture neighborhood dependencies. Pretraining on large collections of synthetic graphs sampled from a rich graph prior exposes the model to wide variation in structure, features, and label mechanisms. OOD generalization is achieved through prior-driven diversity and context-based inference, allowing the model to adapt to new graphs without parameter updates when target distributions fall within the support of the pretraining prior.

GOODFormer~\cite{liao2026invariant} integrates invariant learning into a graph Transformer foundation architecture to support generalization under structural and semantic shifts. It introduces an entropy-guided subgraph disentangler that separates invariant and variant substructures using complementary attention patterns. Predictions are constructed to depend on invariant subgraphs under an invariant learning objective inspired by interventional risk minimization. GOODFormer explicitly controls reliance on distribution-specific patterns by disentangling stable structure-label relations, enabling consistent performance across graphs under distribution shifts.

\subsection{Heterogeneous-Task GFMs for Handling OOD}
Heterogeneous-task GFMs are designed to support flexible task shifts by explicitly modeling variation in the task factor $\Phi_{\text{task}}$ across environments, where supervision forms, output spaces, and inference objectives may change between $p_{\text{src}}$ and $p_{\text{tgt}}$. In this setting, generalization requires mechanisms that remain effective when the distribution shift is coupled with a shift in task specification, so that the same pretrained representation can be generalized under new supervision and evaluation protocols. Existing approaches address this need through task-agnostic pretraining objectives, unified semantic interfaces, and instruction- or prompt-based adaptation that provide a shared control plane for node-, link-, and graph-level use cases. Instead of tying representations to a single label space or fixed head, these models learn structural and semantic primitives that can be composed at inference time to match the target task, which fits open-ended and multi-task deployment where both graphs and tasks evolve.

OFA~\cite{liu2024one} unifies diverse graph tasks by grounding representation learning in text-attributed graphs. Nodes and edges from different domains are described using natural language and encoded into a shared embedding space, enabling graphs from citation networks, molecules, and knowledge bases to be processed by a common encoder. Task heterogeneity is handled through the notion of nodes of interest, which define task-specific subgraphs and introduce a prompt node that aggregates relevant information via message passing. This design converts node-level, link-level, and graph-level problems into a unified classification interface without task-specific readout heads. OFA encodes both structural context and task semantics directly into the graph, allowing generalization when domains, task definitions, or label spaces change. 

LLaGA~\cite{chen2024llaga} builds GFM through alignment between graph structure and large language models. Graphs are reformulated as structure-aware node sequences and projected into the token embedding space of a frozen language model using parameter-free encoding templates and a lightweight projector. This enables a single model to support multiple tasks, including node classification and link prediction, through a shared question--answer interface. The foundation aspect arises from learning a reusable graph-to-token alignment across datasets and tasks. By mapping graphs into a semantic token space shared across domains, this method allows learned structural patterns to be reused on unseen graphs without parameter updates. Empirical results show stable performance under cross-domain evaluation and zero-shot task transfer, highlighting how representation alignment supports task variability.

OpenGraph~\cite{xia2024opengraph} focuses on learning transferable graph representations through universal tokenization and large-scale pretraining. Arbitrary graphs are converted into sequences of topology-informed node tokens using smoothed high-order adjacency information and topology-aware projection, enabling graphs with different sizes and structures to share a common representation space. A scalable graph transformer models global dependencies over these tokens, while masked autoencoding pretraining avoids reliance on task-specific labels. To broaden coverage, OpenGraph incorporates synthetic graphs generated by large language models, increasing structural and semantic diversity during training. These design choices allow the pretrained model to be applied to unseen graphs and tasks in a zero-shot manner, supporting OOD generalization when both graph structure and task formulation differ from training.

GOFA~\cite{kong2025gofa} extends generative language modeling to graphs by unifying textual semantics and relational structure within a single architecture. Graphs are treated as text-attributed, and a generative graph completion objective generalizes next-token prediction to nodes conditioned on local subgraphs. By interleaving graph neural layers with a pretrained decoder-only language model, GOFA enables token-level message passing while preserving a unified generation interface. This supports node-level, link-level, and graph-level reasoning without task-specific heads. Conditioning generation on prompts and subgraphs rather than fixed labels allows the model to generalize to unseen tasks and domains at inference time, enabling zero-shot inference.

LLM-BP~\cite{wang2025generalization} proposes a training-free framework for text-attributed graphs by decoupling text representation from graph aggregation. Node texts are encoded by large language models using task-aware prompting, aligning embeddings with label semantics rather than generic similarity. Structural reasoning is handled through a probabilistic formulation based on belief propagation, where aggregation parameters are estimated via language model queries instead of supervised learning. This combination yields a foundation model that supports inference on unseen graphs and tasks. Generalization arises from avoiding dataset-specific training and from adapting aggregation behavior to different graph connectivity patterns, enabling zero-shot inference across domains and labeling schemes.

GIT~\cite{wangtowards2025} introduces task-trees as a unified abstraction for heterogeneous graph tasks, aligning node-level, edge-level, and graph-level problems within a single representation space. Task-trees connect task-relevant nodes through a virtual task node, forming computation objects compatible with standard message passing. A shared encoder is pretrained using a task-tree reconstruction objective, encouraging representations that capture semantics common across tasks and domains. The design of representing tasks through aligned structures can support OOD generalization when task definitions change, as similar subtrees induce similar embeddings across environments.

GFT~\cite{wang2024gft} proposes a cross-domain and cross-task graph foundation model by learning a discrete vocabulary over computation trees derived from message passing. Pretraining encodes computation trees from large graph collections and learns reusable tree tokens via vector quantization under a tree reconstruction objective that captures feature, topology, and semantic information. At inference, node-, link-, and graph-level tasks are unified as computation tree classification by querying task-specific trees against the fixed vocabulary. This design enables reuse of transferable structural patterns across domains and tasks, improving generalization under distribution shift.

AutoGFM~\cite{chen2025autogfm} addresses task and domain shifts by integrating graph neural architecture search into GFMs. A shared super-network enables weight sharing across datasets, while different architectural patterns are activated based on graph characteristics. Tasks are unified through node-of-interest subgraphs, supporting multiple prediction levels within a common encoder space. It disentangles invariant and variant patterns and only uses invariant information for architecture selection, which reduces sensitivity to dataset-specific variation and enables stable performance under distribution shifts.

UniGraph~\cite{he2025unigraph} uses text as a shared semantic layer to support learning across diverse text-attributed graphs. A cascaded backbone combines a pretrained language model for node text encoding with a graph neural network for structural propagation. Self-supervised masked graph modeling pretraining captures transferable patterns beyond a single task. A unified task formulation based on anchor nodes enables consistent handling of node-level, edge-level, and graph-level predictions. Instruction-based inference further allows prediction using textual category descriptions, enabling adaptation to unseen label spaces. These components together support OOD generalization across graphs, tasks, and domains without retraining.

UniGraph2~\cite{he2025unigraph2} extends this framework to multimodal graphs by unifying structure with heterogeneous node features such as text and images, which is a pioneering work in handling modality-level distribution shifts. Specifically, the modality-specific encoders are combined with graph neural networks, and masked multimodal prediction pretraining captures cross-modal and structural dependencies. Cross-domain multi-graph pretraining and a mixture-of-experts module align representations across domains and modalities. Without task-specific retraining, the learned embeddings transfer to unseen graphs. Evaluations under in-domain and out-of-domain settings show stable and promising performance, indicating that unified embedding spaces and cross-domain alignment support OOD generalization under heterogeneous task and data conditions.

\begin{table*}[t]
\centering
\footnotesize
\caption{Evaluation task categories for graph foundation models under OOD generalization}
\label{tab:evaluation_tasks}
\begin{adjustbox}{max width=0.99\textwidth}
\begin{tabular}{C{0.12\linewidth} M{0.48\linewidth} M{0.52\linewidth}}
\toprule
\multicolumn{1}{c}{\textbf{Category}} &
\multicolumn{1}{c}{\textbf{Representative Task}} &
\multicolumn{1}{c}{\textbf{Evaluation Focus}} \\
\midrule

\multirow{1}{*}{Structural OOD}
& Node / link / graph prediction under topology or feature shift
& Topology and feature statistics shift; connectivity pattern variation \\
\midrule

\multirow{2}{*}{Domain OOD}
& (i) Cross-domain transfer for node / link / graph tasks
& (i) Dataset bias; domain semantics shift; negative transfer avoidance \\
& (ii) Multi-domain generalization and alignment
& (ii) Domain mixture shift; alignment consistency \\
\midrule

\multirow{2}{*}{Modality OOD}
& (i) Node / link / graph prediction with missing modalities
& (i) Modality dependence; missing or corrupted modality generalization \\
& (ii) Multi-modal alignment and grounding evaluation
& (ii) Structure--modality alignment; semantic grounding consistency \\
\midrule

\multirow{2}{*}{Task OOD}
& (i) Unified node / link / graph task generalization
& (i) Task granularity generalization; head-free task reuse \\
& (ii) Instruction- or prompt-driven generalization to unseen tasks
& (ii) Prompt conditioning; unseen task and label-space adaptation \\
\bottomrule
\end{tabular}
\end{adjustbox}
\end{table*}

\section{Evaluation}
\label{sec:evaluation}

Evaluation of GFMs under OOD examines whether pretrained models keep stable performance under distribution shifts. 
Table~\ref{tab:evaluation_tasks} summarizes the main evaluation tasks used in existing studies.

A common evaluation setting measures performance under changes in graph topology or node feature statistics. In these settings, GFMs are trained source graphs and evaluated on target graphs whose degree distributions, motif frequencies, or connectivity patterns differ from those seen during training. Evaluation is typically conducted through cross-graph generalization or by applying structural perturbations, such as edge rewiring or node removal, at test time. These protocols test whether learned representations rely on general structural patterns rather than properties specific to a single graph or dataset. Many traditional graph learning methods report performance drops under such settings, which motivates pretraining on multiple graphs or using structure-aware alignment during training~\cite{li2023invariant,lachi2024graphfm,xia2024anygraph,sun2025riemanngfm}.

Another widely used evaluation category considers generalization across datasets from different domains. In cross-domain generalization settings, models are trained on one or more source domains and evaluated on a testing domain without using testing-domain labels. This setup tests whether models can generalize knowledge across datasets with different data collection processes and label distributions. More complex evaluations consider multiple source domains and evaluate performance when the domain composition changes at test time. These evaluations are often implemented by holding out entire domains during testing. Reported results highlight the impact of dataset bias and negative transfer when models rely too strongly on dominant domains~\cite{lin2025unified,he2025unigraph2}.

Evaluation under modality variation focuses on settings where auxiliary modalities, such as text or images, are missing or corrupted at test time. One group of protocols measures how performance changes when certain modalities are removed, which reflects incomplete data in real applications. Another group evaluates whether representations learned from graph structure remain consistent with auxiliary modality representations across datasets. These evaluations test whether models depend heavily on specific modalities or can maintain performance when modality availability changes. Prior work commonly implements such evaluations through modality masking or cross-dataset transfer with different modality configurations~\cite{cai2024multimodal,he2025unigraph2,wang2025generalization}.

We also noticed that several studies evaluate generalization across task specifications. One evaluation setting considers transfer across node-, link-, and graph-level tasks using a single pretrained model, without retraining task-specific components. This setup tests whether a shared representation can support multiple prediction types. Another setting evaluates instruction- or prompt-based inference, where task descriptions and output formats are provided at inference time. These evaluations are usually conducted in zero-shot or few-shot settings and assess whether models can adapt to new tasks without additional training~\cite{liu2024one,chen2024llaga,wangtowards2025}.

Overall, the evaluation tasks summarized in Table~\ref{tab:evaluation_tasks} show that out-of-distribution assessment for GFMs covers multiple dimensions. Different evaluation protocols test different aspects of generalization, and no single protocol is sufficient on its own. Studies that combine evaluations across structure, domain, modality, and task provide a more complete evaluation of model performance under distribution shift.

\section{Future Directions}
\label{sec:future}
Despite rapid progress of GFMs for OOD generalization, there still exist plenty of opportunities worthy of future explorations.

\begin{itemize}[leftmargin = 0.3cm]
    \item \textbf{Toward a universal graph vocabulary.}  
    A central obstacle for GFMs is the absence of transferable atomic units analogous to tokens or patches~\cite{dosovitskiy2021an,yu2025graph2text}. Existing attempts based on computation trees, geometric primitives, or text proxies remain domain-dependent. Future work could explore what constitutes a generalizable graph unit, and how such units can preserve domain-specific semantics while enabling foundation-level pretraining and OOD generalization.

    \item \textbf{Establishing scaling laws for graph models.}  
    It remains unclear whether predictable scaling relationships between data, model capacity, and performance exist for graphs. It requires deeper study of how graph size, dataset diversity, and architectural capacity interact, as well as the design of unified generative or reconstruction objectives that can support meaningful scaling behavior~\cite{liu2024towards}.

    \item \textbf{Aligning pretraining with OOD objectives.}  
    Existing works show that naive graph pretraining can induce negative transfer under distribution shift~\cite{li2024disentangled}. Therefore, an important direction is to understand when and why pretraining helps or harms OOD generalization, and to develop objectives that explicitly favor stable, generalizable mechanisms rather than domain-specific structural correlations.
    
    \item \textbf{Theoretical guarantees under graph distribution shifts.}  
    Existing theoretical guarantees for GFMs are limited and largely rely on independent and identically distributed (IID) assumptions that do not align with foundation-level pretraining. Future theory should aim to characterize OOD generalization in GFMs by modeling how pretrained representations generalize across environments, while accounting for structural variation, feature heterogeneity, and graph size diversity inherent to large-scale graph collections~\cite{morris2024future}.    
    
    \item \textbf{Benchmarking realistic OOD deployment scenarios.}  
    There are a lot of benchmarks and evaluation protocols introduced for GFMs and OOD generalization~\cite{gui2022good,xu2025graphomni}. Nevertheless, it is still important to develop standardized benchmarks more systematically that capture realistic structural, temporal, multimodal, and domain shifts, support consistent evaluation across methods, and enable statistically reliable comparisons. 

\end{itemize}

\medskip
{
\small
\bibliographystyle{unsrt}
\bibliography{output}
}

\end{document}